\RequirePackage{fix-cm}
\documentclass[smallcondensed]{svjour3}     
\smartqed  

\usepackage{lineno,hyperref}
\usepackage{graphicx}
\usepackage[export]{adjustbox}
\usepackage{amsmath}
\usepackage{amsfonts}
\usepackage{amssymb}
\usepackage{hyperref}
\usepackage{booktabs}
\journalname{Springer Nature Artificial Intelligence Review}
\begin{document}

\title{Neural Attention for Image Captioning: Review of Outstanding Methods 
}


\author{Zanyar Zohourianshahzadi*  \and
        Jugal K. Kalita 
}


\institute{* Corresponding Author: Zanyar Zohourianshahzadi \at
              Department of Computer Science, University of Colorado Colorado Springs, 1420 Austin Bluffs Pkwy, Colorado Springs, Colorado, CO 80918 \\
              Tel.: +17194245976 \\
              \email{zzohouri@uccs.edu}           
           \and
           Jugal K. Kalita \at
           Department of Computer Science, University of Colorado Colorado Springs, 1420 Austin Bluffs Pkwy, Colorado Springs, Colorado, CO 80918\\
           \email{jkalita@uccs.edu}
}

\date{Received: date / Accepted: date}
\section*{Declarations}
\subsection{Funding}
This work has not received any funding. N/A

\subsection{Conflicts of interest/Competing interests}
The authors declare that they have no conflict of interest or competing interest in this work. N/A

\subsection{Availability of data and material (data transparency)}
The data-set used in all experiments and results reported in this work is publicly available for download. COCO data-set is available at ``cocodataset.org''.

\subsection{Code availability (software application or custom code)}
The code for some of the papers reviewed in this work were originally published by the authors on github. The reader of this work can 
lookup for the code for a particular paper reviewed in this work online.
a
\subsection{Authors' contributions}
Zanyar Zohourianshahzadi collected all the research material and wrote the original text of the survey paper. Jugal K. Kalita helped with article
preparation phase and provided suggestions regarding the text and structure of the paper.
\maketitle

\begin{abstract}
        Image captioning is the task of automatically generating sentences that describe an input image in the best way possible.
        The most successful techniques for automatically generating image captions have recently used attentive deep learning models.
        There are variations in the way deep learning models with attention are designed.
        In this survey, we provide a review of literature related to attentive deep learning models for image captioning. 
        Instead of offering a comprehensive review of all prior work on deep image captioning models,
        we explain various types of attention mechanisms used for the task of image captioning in deep learning models.
        The most successful deep learning models used for image captioning follow the encoder-decoder architecture,
        although there are differences in the way these models employ attention mechanisms.
        Via analysis on performance results from different attentive deep models for image captioning, we
        aim at finding the most successful types of attention mechanisms in deep models for image captioning.
        Soft attention, bottom-up attention, and multi-head attention are the types of
        attention mechanism widely used in state-of-the-art attentive deep learning models for image captioning.
        At the current time, the best results are achieved from variants of multi-head attention with bottom-up attention.
\end{abstract}

\newpage
\section{Introduction}\label{sec1}
\par
Image captioning creates a nexus between natural language processing and computer vision. 
Image captioning systems could be used for a variety of tasks.
Although such systems have been mostly trained on libraries of mundane images such as ones obtained by mining the web, one can easily
think of beneficial use cases.
For instance, these systems could enable blind individuals to receive
visual information about their surrounding environment.
When these systems are trained on medical images, they could be used
to provide physicians with useful information and help with the diagnosis procedures \cite{Pavlopoulos_2019_survey}.
Other applications of image captioning include image-sentence search and retrieval as well as bringing visual intelligence
for robots. Automatic image captioning can also be used in advanced recommendation and visual assistant systems.
\par
Template generation and slot filling \cite{Farhadi_2010,Kulkarni,Li_2011}
and caption retrieval \cite{Ordonez_NIPS2011,Gong_2014,Hodosh_2015,Sun_2015_ICCV} are
the early techniques used for automatic image caption generation.
In comparison with template-based and retrieval methods, better results have been achieved by employing deep neural networks \cite{kiros2014_1,Kiros2014_2,Karpathy_2014,Karpathy_2015_CVPR}.
The common deep learning architecture used
is the encoder-decoder \cite{Sutskever_2014,Encoder_Decoder_Cho,kalchbrenner} architecture.
The encoder-decoder scheme was introduced in the context of neural machine translation.
Using the encoder-decoder approach, we can divide
the translation task into two parts, the encoding phase extracts features from input data,
and decoding phase creates the output with respect to the encoded or extracted features.
Deep learning methods learn visual features
by employing convolutional neural networks \cite{Resnet,VGG} (CNNs) or a combination of object detectors \cite{RCNN,FastRCNN,FasterRCNN} with CNNs.
In image captioning, we use the encoder-decoder architecture like how they are used in neural machine translation,
since we map visual features to a sequence of tokens, analogous to mapping an input sequence of words to an output sequence for translation.
The tokens are the words of a sentence in array before they are used in
the process of generating word embeddings.
Early deep attentive image captioning models employed convolutional neural networks (CNN) as encoders and
long short-term memory networks (LSTM) \cite{LSTM} as decoders \cite{Xu,Cho}.
In most of the work published recently and reviewed in this survey,
bottom-up attention \cite{Anderson2017up-down} is used for visual feature extraction, which we explain in Section \ref{sec4}.
\par
Image captioning using deep learning usually involves supervised learning. Although recently, Laina et al. \cite{Laina_2019_ICCV}
and Feng et al. \cite{Feng_2019_CVPR}
have showed that training a deep learning model for image captioning could lead to desirable results using unsupervised learning,
the best results still come from the models trained with supervised learning.
Therefore, it is not necessary to categorize the papers reviewed for this
survey under supervised learning and unsupervised learning.
In supervised learning, we always train the model
with a dataset containing examples labeled with the ground truth of the output.
\par
\begin{figure}[t!]
    \centering
    \includegraphics[width=\linewidth, frame={\fboxrule} {-\fboxrule}]{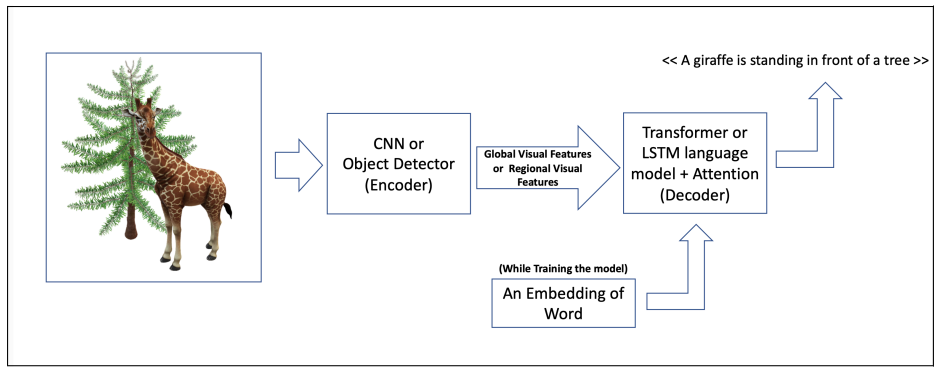}
    \caption{The task of image captioning using attentive deep learning models employing encoder-decoder architectures. The majority of
    deep learning models for image captioning use the encoder-decoder architecture.}
    \label{fig1}
\end{figure}
\par
Different types of encoders feed different types of information to the attention mechanisms and language models (decoders).
After performing a review of the evolution of attention mechanisms in image captioning,
we categorize the state-of-the-art literature based on the types of attention mechanisms used alongside different kinds of encoders such as CNNs or CNNs and object detectors (bottom-up attention encoders)
or a combination of these with graph convolutional networks (GCN) \cite{GCN_Bruna,GCN_Defferrand}, coupled with various types of decoders, primarily LSTMs \cite{LSTM} and Transformer \cite{NIPS2017_7181}.
The Transformer is an encoder-decoder based model that consists of Multi-Head Attention (MHA) and Scaled-dot Attention (Self-Attention).
Our goal in this survey is to review and reveal the best practices in employing attention mechanisms for
image captioning in deep neural networks, specifically among the state-of-the-art methods that achieve better performance in comparison with earlier types
of attention mechanisms such as spatial soft and hard attention or semantic attention.
\par
In Section \ref{sec2}, we provide a brief review of previous
surveys of deep learning models used for automatic image caption generation. In Section \ref{sec4}, we discuss the evolution path of attention mechanisms
from earlier methods and how they inspired state-of-the-art literature. In Section \ref{sec4.3}, we offer a performance comparison
among the state-of-the-art methods reviewed in our survey.
\par

\section{Related Work}\label{sec2}
\par
Recently several high-quality surveys on image captioning with deep neural networks
\cite{Hossain:2019:CSD:3303862.3295748,Liu2019_survey} and
other techniques such as template generation and slot filling \cite{BAI_AN_2018_survey,Sharma_2020_survey,Li_2019_Survey} have been published. 
Instead of reviewing all methods used for image captioning, we record the success
the use of attention mechanisms has brought to image captioning.
After briefly reviewing the history of attention mechanisms in image captioning, we review
the state-of-the-art literature to highlight the most successful attention mechanisms.
This is what mainly differentiates our work from previous surveys on deep learning models for image captioning.
\par
In general, researchers have recently noticed the usefulness of self-attention and multi-head attention over recurrent neural networks.
The Transformer \cite{NIPS2017_7181} model utilizes
self-attention (scaled dot-attention) inside multi-head attention to obviate the use of recurrences
and convolutions, relying fully on attention for neural machine translation.
We provide a brief review of the methods that use the Transformer
model or its variants for image captioning as well as
other methods that use LSTMs with attention mechanisms.
\par
A comprehensive survey of deep learning models for image
captioning by Hossain et. al. \cite{Hossain:2019:CSD:3303862.3295748}, creates a well-defined taxonomy for
the models. However, the most successful methods
published after this paper have used
Transformers \cite{NIPS2017_7181} or variations of this model. The same applies
to the survey by Liu et al. \cite{Liu2019_survey}. Although these surveys provide good background on evaluation metrics,
datasets and methods used for automatic image caption generation, for the most part
they do not discuss the subtle differences among the attention mechanisms used in
the deep learning models for image captioning.
This is mainly because the methods published after these surveys used bottom-up attention encoders and Transformer
or components of this model such as multi-head attention, which improves performance by adding more attention units in a parallel manner.
Multi-head attention was introduced by Vaswani et al. \cite{NIPS2017_7181}. 
\par
Another survey by Bai and An \cite{BAI_AN_2018_survey},
reviews methods that use deep learning as well as other non-empirical methods.
This survey and other similar surveys performed by Sharma et al. \cite{Sharma_2020_survey} and Li et al. \cite{Li_2019_Survey}
on automatic image caption generation do not perform a review of attentive deep learning models for image captioning.
\par
A noteworthy survey by Pavlopoulos et al. \cite{Pavlopoulos_2019_survey},
reviews, deep learning models and other techniques used for medical image captioning.
Given a large dataset of medical images and relevant diagnosis sentences for each image,
it is possible to train deep learning models used for image captioning to generate candidate diagnosis sentences.
This is likely to help physicians and save them considerable time when performing diagnosis procedures using visual medical data. 
This survey also reviews the datasets used for medical image captioning.
\par
The comprehensive survey performed by Hossain et al. \cite{Hossain:2019:CSD:3303862.3295748}, offers
useful information regarding common datasets, metrics
and categories for deep learning models used for image captioning.
We avoid repeating the same information
and instead focus on the state-of-the-art models that use attention mechanisms. 
We also emphasize work that has been published after the previous surveys
and offer a new taxonomy for attentive deep learning models used for image captioning.
\par

\section{Attentive Deep Learning for Image Captioning}\label{sec4}
\par
In this section, first we discuss the early use of attention mechanisms in image captioning deep models.
We follow by the introduction of bottom-up and top-down attention \cite{Anderson2017up-down} (Up-Down Attention),
which became a source of inspiration for most of the later work.
In recent years, the use of generative adversarial networks (GANs) for image captioning has also led to good results \cite{Dai_2017_ICCV,Chen_GAN,Liu_Dual_GAN}.
In comparison with the encoder-decoder architecture, which is usually trained with cross-entropy loss,
GAN architectures are trained with adversarial loss, making it impossible to perform a direct comparison of performance.
Although recently attention mechanism was employed inside a GAN model \cite{WEI_GAN_2020}, this model utilizes the encoder-decoder architecture
inside the generator and discriminator modules in order to make use of attention. Considering that attention mechanisms
have been widely used in encoder-decoder architectures, we present the way attention is calculated and used in these architectures.
\par
\begin{figure}[ht!]
    \centering
    \includegraphics[width=\linewidth, frame={\fboxrule} {-\fboxrule}]{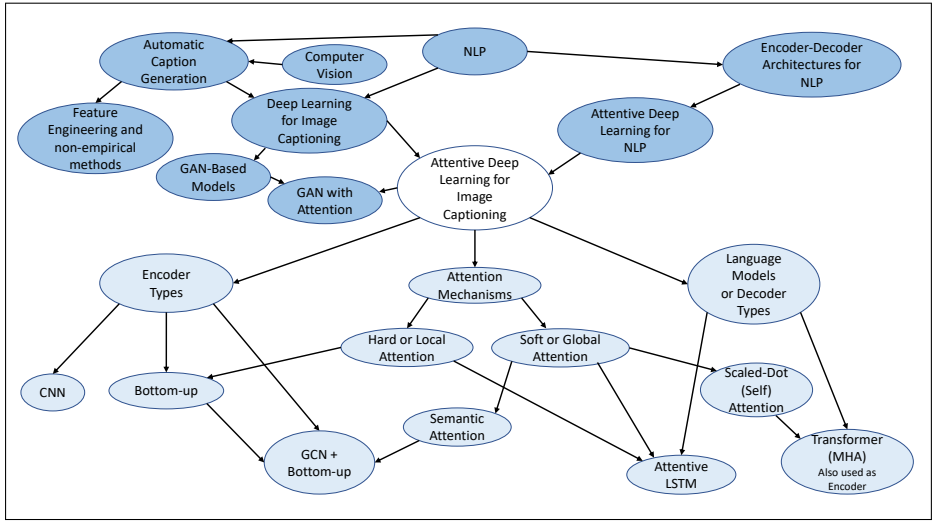}
    \caption{A taxonomy graph of attentive deep learning models used for image caption generation in our survey.
    Attentive deep learning for image captioning
    combines computer vision, encoder-decoder architecture and attention mechanism.
    In our survey, we focus on encoder and decoder types as well as attention mechanisms that are used
    in the state-of-the-art methods employed for attentive deep image captioning.
    Bottom-up Attention: Object-Detector + CNN Backbone -
    MHA: Multi-Head Attention (utilizes Self-Attention) -
    GCN: Graph Convolutional Networks -
    LSTM: Long Short-term Memory -
    Transformer: Employs MHA and Scaled-dot (Self) Attention -
    NLP: Natural Language Processing.}
    \label{fig2}
\end{figure}

After a review of history of attention mechanisms used for image captioning in the context of encoder-decoder architecture,
we elaborate upon the state-of-the-art attention mechanisms in the context of different kinds of encoders in which they were used.
Different types of encoders provide the attention mechanisms and paired decoders with different kinds of input information.
Therefore, it is necessary to analyze the differences among attention mechanisms in the context of the associated encoders.
We have referred to the deep learning models that use attention mechanisms as attentive deep learning models.
A taxonomy graph of technologies associated with attentive deep image captioning is illustrated in Fig. \ref{fig2}. In our survey, we focus
on attention mechanisms in the context of state-of-the-art encoder-decoder architectures for image captioning.
\par
\subsection{Evolution path of attention for image captioning} \label{sec4.0}
\par
We have witnessed the emergence and widespread use of attention
mechanisms in deep learning models during the past few years.
Attention mechanisms have a long history in neuroscience in the context of visual attention, dating back a few decades \cite{Koch1987,DEUBEL1996}. 
A few years ago, attention mechanisms started to show their usefulness in deep learning models in the
context of neural machine translation \cite{ORG_Attention}. This led researchers to investigate
the usefulness of attention mechanisms in image captioning \cite{Xu,Cho}.
\begin{figure}[t!]
    \centering
    \includegraphics[width=\linewidth, frame={\fboxrule} {-\fboxrule}]{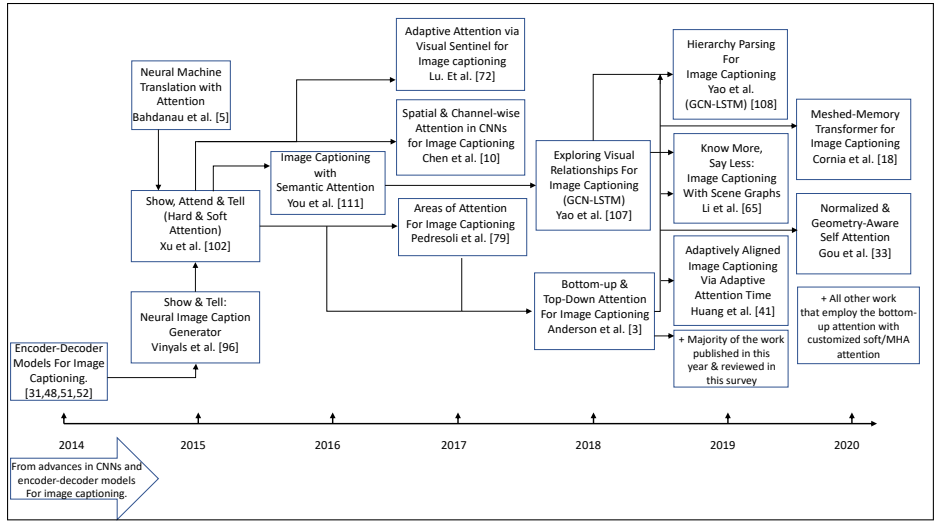}
    \caption{The timeline of advent of seminal works in attentive deep image captioning.
    The majority of state-of-the-art attentive methods are inspired by soft and hard attention in Show, Attend and Tell \cite{Xu}
    and Bottom-up \& Top-Down (Up-Down) Attention \cite{Anderson2017up-down}. Each block shows the title of work and author name. For more details refer to text.}
    \label{fig3.1}
\end{figure}
Fig. \ref{fig3.1} shows the seminal works in attentive deep image captioning.
In this figure, starting from the left we show the pioneering works in CNNs and encoder-decoder models for image captioning.
\par
CNNs have shown their effectiveness in visual feature extraction in image classification \cite{CNNLecunNet1,Lecun_Bengio,CNNOriginal,AlexNet}.
The pixels in an image have long range
dependencies among one another when forming patterns, and therefore CNNs are useful in
visual feature extraction for image captioning \cite{Vinyals_2015_CVPR,Donahue_2015_CVPR}.
After training a CNN on image classification task on a huge dataset such as the ImageNet \cite{imagenet_cvpr09},
we remove the last SoftMax feed-forward layer used for image classification.
This mechanism provides us with a CNN backbone that can be used for feature extraction from unseen images. 
The CNN backbone used for feature extraction among majority of state-of-the-art attentive deep learning models for image captioning is ResNet101 \cite{Resnet}.
\par
The first attentive deep model for image captioning was Show, Attend and Tell \cite{Xu}.
This model used a CNN namely the VGG model \cite{VGG} pre-trained on ImageNet \cite{imagenet_cvpr09}
for feature extraction, and LSTM
for language modeling as the decoder.
This was very similar to other encoder-decoder (CNN-LSTM) architectures for image captioning \cite{Vinyals_2015_CVPR,Donahue_2015_CVPR,Karpathy_2015_CVPR},
except that Show, Attend and Tell \cite{Xu} employed two variants of attention mechanisms,
namely soft and hard attention on the spatial convolutional features to provide a set of attended features
for the LSTM decoder that acts as a language model.
\par
In Fig. \ref{fig3.1}, we do not show the advent of Transformer \cite{NIPS2017_7181} and Resnet \cite{Resnet}
since it would increase the complexity of this diagram. It is important to note that Up-Down attention model
utilizes Faster-RCNN \cite{FasterRCNN} with ResNet101\cite{Resnet} as bottom-up attention
for visual feature extraction and high-performing state-of-the-art models have used the bottom-up attention with
multi-head attention in Transformer model for image captioning rather than using the traditional attentive LSTM decoder.
\par
\begin{figure}[ht!]
    \centering
    \includegraphics[width=\linewidth, frame={\fboxrule} {-\fboxrule}]{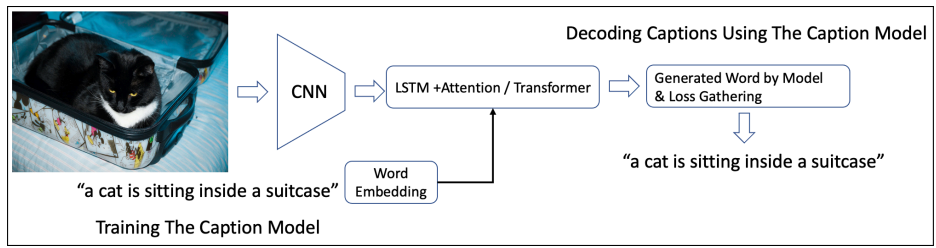}
    \caption{Attentive decoders with CNN encoder for feature extraction from a general point of view.}
    \label{fig3}
\end{figure}
\par
In this section, we present work that introduced attention mechanisms as well as other work that became a source of inspiration for
most recent deep attentive models for
image captioning until the advent of Up-Down attention model. We finish this section by introducing the Up-Down attention model and then
start the next section by exploring state-of-the-art literature categorized based on the kinds of encoders and decoders within which attention is deployed.
\par
In Fig. \ref{fig3} we show the models that utilize LSTM with attention mechanism on extracted visual features as well as models that use
Transformer encoder on extracted visual features from a CNN backbone, and Transformer decoder that receives
a masked embedding of the ground truth caption sequence while training. This figure
can be applied to all the methods we explain in Section \ref{sec4.0} and Section \ref{sec4.1.1} from a general viewpoint.
\par
\par
\subsubsection{Soft \& Hard Attention}
\par
In the Show and Tell model introduced by Vinyals et al. \cite{Vinyals_2015_CVPR}, the extracted visual features from a CNN enhanced with batch normalization \cite{Batch_Norm_2015}
are passed to an LSTM network for hidden state initialization, whereas in the Show, Attend and Tell model introduced by Xu et al. \cite{Xu},
the set of attended extracted visual features are concatenated with an embedding of each word in the caption sentence at each time step
to form the input for the decoder.
Applying attention mechanism on visual features and passing the attended visual features to the LSTM decoder would enable it
to attend over particular regions in the input image and the word embeddings concurrently. Both models use a feed-forward layer as an embedding layer for a one-hot vector representing
the words in the caption sentence from a vocabulary.
\par
Word embeddings are vector representations of the tokens that are fed to a deep learning model. Training simultaneously on word embeddings
and attended visual features creates the ability to learn
patterns for the words in a sentence and visual features in a shared embedded space.
The most common embedding systems used for natural language processing
and image captioning are Glove \cite{Glove} and Word2vec \cite{Word2vec}.
An easier approach used sometimes is a one-hot vector of the current word in the sentence
as input with a size equal to the length of vocabulary. An embedding (feed-forward) layer is then used to encode the one-hot vector into word embeddings.
Recurrent neural networks such as
long short-term memory networks (LSTMs) \cite{LSTM} and gated recurrent units (GRUs) \cite{GRU} as well as Transformer \cite{NIPS2017_7181}
can be used for mapping embedded words to visual features for image caption generation.
\par
The LSTM decoder with attention in Show, Attend and Tell receives the set of attended visual features ($\hat{z}$) from an attention function ($f_{att}$),
which may either be a soft deterministic or a hard and stochastic
function. 
The LSTM decoder also receives $E(y_{t-1})$, the word embedding of the previous word in the caption sentence
at each time step, and the previously generated hidden state ($h_{t-1}$)
created by the LSTM in order to create the next hidden state used for generating the next word at time step $t$.
\par
The attention values ($e_{t}$) are created by the attention function ($f_{att}$), which is a linear transformation layer that
receives the previous state ($h_{t-1}$) and the extracted visual feature vectors ($\textbf{a}$), where each annotation vector ($\textbf{a}_i$)
is a spatial feature vector referring to a specific part of the input image.
The final soft weighting vector ($\alpha_{t}$) is created from application of SoftMax function over the attention values ($e_{t}$). 
After the soft attention weights ($\alpha$) are computed we can apply them to the set of extracted spatial visual
features in soft attention function ($\phi_{soft}$). At each time step, the soft attention weights ($\alpha$)
are multiplied by the feature vectors ($\textbf{a}$).
\par
The hard attention function ($\phi_{hard}$) 
employs a one-hot vector $s_t$ parametrized by the weighting vector ($\alpha$).
At each time step $t$ the i-th location in the one-hot vector ($s_{ti}$) is set to one if the i-th location
in the image is the one being used for feature extraction.
This way the model only attends over one annotation vector ($a_i$) rather than a group of weighted annotations as in soft attention.
\par
Both soft and hard attention may be used to calculate
the set of attended visual features ($\hat{z}$) for use in an attentive LSTM for generating the hidden state
used for caption word prediction. 
For word generation at each time step by the model the hidden state is calculated by the LSTM decoder in the model, which receives a
concatenation of attended visual features ($\hat{z}$) with the hidden state vector generated at previous time step ($h_{t-1}$)
and an embedding of the previously generated caption word ($Ey_{t-1}$).
\par
\par
It is important to note that both soft and hard attention mechanisms rely on spatial features extracted from the 2-dimensional image.
Considering that an image is represented using three color channels (Red, Blue and Green),
the extracted annotation vector contains features extracted from each
color channel in a 3-dimensional spatial feature vector.
After the set of attended spatial features is calculated, they are ready for use in the attentive LSTM
for calculating the next hidden state.
The optimization techniques used in Show, Attend and Tell were RMSprop \cite{RMSProp} in
Flickr30k \cite{flickr30k}, and ADAM \cite{Adam} in MS-COCO \cite{MS_COCO}.
We suggest that the readers refer to published literature cited in this work for additional technical details such
as how the objective functions for soft and hard attention differ or how the one-hot vector $(s_t)$
for hard attention is calculated using a Multinoulli distribution.
\par
The use of attention mechanism greatly improved the performance of the Show, Attend and Tell model compared to Show and Tell.
The soft visual attention mechanism that was explained here was used for describing video
and image content and translation by Cho et al. \cite{Famous_Cho_etal}
in a similar way it was used here except that the attention values from current time
step are sent to the attention function for calculating the attention values for the next time step.
Yang et al. \cite{Yang_2016_CVPR} and Xu et al. \cite{Xu2} created stacked
spatial attention models in order to improve soft attention for visual question answering (VQA).
In these models, the stacked attention values are calculated based on attended extracted features modulated by the lower-level attention values.
Soft and hard visual attention have been a source of inspiration for a most of the state-of-the-art literature.
The downside of using spatial features from the whole image for attention is that fine-grain details from objects and their correlations
are not considered. 
\par
\subsubsection{Semantic Attention}
Using semantic information for image captioning with encoder-decoder architecture was explored earlier \cite{Jia_2015_ICCV}.
Introduced by You et al. \cite{You_2016_CVPR}, the main idea behind the introduction of semantic attention was to find a way to
consider semantic attributes discovered in the input image for attention calculation.
Various techniques can be used for attribute detection such as using weakly annotated images on the web with hash tags and captions.
However, the best method for attribute detection turned out to be using a Fully Convolutional Network (FCN) \cite{Long_2015_CVPR}, like
the work by Fang et al. \cite{Fang_2015_CVPR}.
\par
The attentive LSTM with semantic attention receives soft-weighted embeddings of attributes for input and output attention models.
Inspired by soft attention, the semantic attention creates a set of soft-weighted attention
values for the input attention model ($\phi$)
by incorporating the embeddings of extracted semantic attributes with an embedding of previously generated word by the model.
\par
The set of extracted visual features ($v$) is provided by the last convolutional layer of the GoogleNet model \cite{GoogleNet}.
Like Show and Tell \cite{Vinyals_2015_CVPR}, the extracted visual features are fed to an LSTM for hidden state initialization.
The extracted visual features are passed through a linear transformation layer.
The output attention model ($\varphi$)
receives the hidden state generated by the attentive LSTM shown as well as the embeddings of extracted semantic attributes.
\par
The input attention model ($\phi$) creates the input $(\textbf{x}_t)$ for the LSTM at each time step. The input attention model
applies a set of attention values $(\alpha)$ to the set of attributes $(y)$ in a soft fashion like soft attention.
The set of attention values $(\alpha)$ is calculated using a bilinear function over one-hot representations of previously predicted words and
attributes. The attention values are then used for normalizing attributes in a SoftMax fashion.
A weight matrix is applied to the sum of embeddings of attributes soft-weighted 
by the attention values that are projected using a diagonal matrix
and added to the embedding of previously predicted word.
\par
The output attention model ($\varphi$) creates the output of the model $(p_t)$ based
on the exponent of a projection of the current hidden state added to the
sum of attention values $(\beta)$ multiplied by the embeddings of attributes
passed through a sigmoid gate and projected by a diagonal matrix.
The attention values $(\beta)$ are calculated by multiplying the current hidden
state multiplied by a linear transformation (projection) of embedding of
each attribute passed through a sigmoid gate.
\par
Semantic attention has showed better performance compared to visual spatial soft and hard attention.
Inspired by semantic attention, Yao et al. \cite{Yao_boosting_2017} employed attributes as semantic information to model the attention
to the locally previous words instead of using attributes as complementary representations.
Instead of using discovered attributes as semantic information, Mun et al. \cite {Mun_2017} introduced the text-guided attention
model, where a guidance caption is randomly selected among relevant captions
gathered from similar images that were found using the nearest neighbor algorithm.
Text-guided and semantic attention are similar in that they both calculate attention with respect to textual information related to the image.
They are also different in that text-guided attention uses captions directly as source of attention for guiding the visual attention.
Zhou et al. \cite{Zhou_TC_2017} introduced text-conditional attention, which models text-conditional features
as a text-based mask on image features. This way text-conditional features are considered as semantic information.
Inspired by the soft attention and semantic attention, Chen et al. \cite{Chen_2017_CVPR} introduced the spatial and channel-wise attention, where
channel-wise attention resembled the semantic attention regarding what should be looked at without discovering the attributes and instead
utilized spatial features.
The downside of semantic attention is that the quality of generated
captions highly relies on the quality of detected attributes from the input image.
Additionally discovering attributes from the image requires more external resources, which leads to increased complexity of the model.
\par
\par
\subsubsection{Spatial \& Channel-wise Attention}
\par
Both soft and hard attention in Show, Attend and Tell \cite{Xu} operate on spatial features.
In spatial and channel-wise attention (SCA-CNN) model,
channel-wise attention resembles semantic attention because each filter kernel in a convolutional layer acts as
a semantic detector \cite{Chen_2017_CVPR}.
In SCA-CNN \cite{Chen_2017_CVPR}, the caption word $(y_t)$ is generated by
mapping the probability vector $(p_t)$ to a dictionary. The probability vector $(p_t)$ is calculated by a applying a SoftMax activation
on the hidden state generated by the attentive LSTM $(h_t)$ and an embedding of the previous word generated by the model $(y_{t-1})$.
The hidden state at each time step ($h_t$) is calculated by the LSTM which receives the concatenation of generated hidden state at previous step and
spatial and channel-wise attended visual features $(X^L)$ and an embedding of the previous word generated by the model $(y_{t-1})$.
\par
The attended features $(X^L)$ are gathered across all convolutional layers, where $L$ is the total number of layers.
The spatial and channel-wise attended visual features $(X^l)$ from each layer $l$ are calculated by applying a modulation function $(f)$,
which is an element-wise multiplication function between extracted attended features $CNN(X^{l-1})$ from the previous convolutional layer
and current attention values $(\alpha,\beta)$ for spatial and channel-wise attention mechanisms $(\Phi_s,\Phi_c)$.
\par
The visual features $(V)$
for spatial attention are reshaped by flattening width and height, where each visual feature vector $(V_i)$ is the visual feature of the i-th location.
The visual features are also reshaped before input to channel-wise attention. Specifically, each vector $(v_i)$ in reshaped features
is the i-th channel in visual features $(V)$ and the length of the reshaped features is equal to the number of channels.
Considering the reshaped features $(V,v)$ for spatial and channel-wise attention,
the soft attention values for spatial ($\alpha$) and channel-wise attention $(\beta)$ are calculated by applying a SoftMax function over the
linear transformation of attention values that are calculated via applying a Tanh function over the result of element-wise addition between
linear transformations of reshaped features and hidden state generated at previous time step.
\par
Considering spatial and channel-wise attention, we can calculate the attended features in two different ways.
We can either calculate spatial attention and use it to modulate the visual features for use in channel-wise attention
(spatial-channel) 
or we can calculate the channel-wise attention and use for modulating the visual features for spatial attention (channel-spatial).
\par
In SCA-CNN, the best results were achieved when channel-wise attention was used for modulating spatial attention inside two layers of ResNet101 (res5c \& branch2b).
Spatial and channel-wise attention showed better performance compared to Show, Attend and Tell and semantic attention models.
Spatial and channel-wise attention are powerful since they can guide the LSTM to where to look at (spatial) and what to look at (channel-wise).
The idea of when to look at the visual features and when to rely on textual
information for caption word prediction was introduced by Lu et al. \cite{Lu2017KnowingWT},
which we discuss next. Like Show, Attend and Tell, all these models rely on spatial visual features directly without considering
the object level details and relationships.
\par
\par
\subsubsection{Adaptive Attention}
\par
As mentioned earlier, the idea behind adaptive attention is to find a way for the model to know when it should focus on visual features and
when it should focus on textual features for caption generation.
Unlike Show, Attend and Tell, instead of passing the attended visual features calculated based on the previous hidden state to an
LSTM for caption word generation at each time step, 
adaptive attention receives the current hidden state and the adaptive context vector to be passed through a non-linear function for
caption word prediction.
\par
The adaptive context vector ($\hat{c}_t$) is calculated based on the attended spatial feature context vector ($c_t$).
The attended spatial context vector is calculated via soft-weighting applied to visual features ($V = [v_1,v_2,...,v_k]$) modulated by attention values
that are obtained via applying a SoftMax function over 
the result of linear transformations of the set of visual features and the hidden state of LSTM added together and passed through a nonlinear transformation layer.
\par
The visual sentinel ($s_t$) is achieved by applying a sentinel gate ($g_t$) to the memory of LSTM ($m_t$) passed through a Tanh function.
The sentinel gate ($g_t$) is achieved via applying a sigmoid function over the linear transformations of concatenation of the embedding of the
previously generated word ($Ey_{t-1}$) and global features vector ($V$) added to the linear transformation of previously generated hidden state ($h_t$)
After the visual sentinel is calculated based on the sentinel gate it is ready to be included in adaptive context vector.
\par
A new sentinel gate ($\beta$) is included in the adaptive context vector,
which is a scalar between 0 and 1. The final sentinel gate ($\beta$) decides whether the spatial features context vector should be used,
or the visual sentinel ($s_t$) should be used to form the adaptive context vector. The final probability over the words in the dictionary is calculated based on a SoftMax feed-forward layer
applied to adaptive context vector added to the hidden state vector.
\par
Adaptive attention showed competitive performance compared to other types of attention mechanisms. Like semantic and visual soft and hard attention mechanisms,
adaptive attention does not consider object-level details making it hard to know if the model is actually looking at the correct object in the image
for word prediction.
Inspired by adaptive attention and Up-Down attention, Lu et al. \cite{Lu2018NBT} introduced the NBT model that could generate captions, which include
words that are visually grounded to particular regions in the image. The visual sentinel in adaptive attention is used in NBT and then
two pointer vectors \cite{Oriol_2015_Ptr} are used for category and subcategory detection for each word that is grounded in the image.
\par
\subsubsection{Bottom-up \& Top-down Attention}
\par
The idea of incorporating regional features for attention mechanism was explored before the advent of Up-Down attention.
Jin et al. \cite{Jin_align} discover the regions related to objects in the image using a selective search technique \cite{Selective}.
The regions are filtered with a classifier, and then resized and encoded using a CNN for input to a soft attention mechanism for image captioning.
In Areas of Attention model, Pedersoli et al. \cite{Pedersoli_2017_ICCV} use spatial transformers \cite{Spatial_trans} and
edge boxes \cite{Objects_edges} for detecting important regions related to objects in the image for feature extraction.
In Global-Local Attention model \cite{Li_global_local}, like in Up-Down attention model, Faster-RCNN \cite{FasterRCNN} is used for detecting salient regions
in the image for feature extraction, where the global features are extracted from the image as a whole and used in a soft attention mechanism
along with local features extracted from object detection regions.
\par
In Up-Down attention model \cite{Anderson2017up-down}, global features
are not considered and instead a mean-pooled average of features from each detected region (bounding box) is considered for the Top-Down soft attention mechanism.
A non-maximum suppression for each object class using an IoU threshold is performed.
Regions with class detection probability that exceeds certain confidence threshold are selected for feature extraction.
Employing Faster-RCNN in this fashion creates a Bottom-up hard attention mechanism as only a small number of bounding box detections are considered for feature extraction.
In Up-Down attention model, Resnet-101 \cite{Resnet} is used in conjunction with Faster-RCNN for feature extraction from bounding box detections.
\par
The captioning model in the Up-Down attention model consists of two LSTMs. The first LSTM acts as a top-down visual attention model, where the hidden state
of the first LSTM is passed to the attention mechanism for generating the set of attended visual features ($\hat{v}$) calculated based on the set of visual
features extracted by bottom-up attention.
The input for the attention LSTM contains the mean-pooled image features ($\bar{v}$) from each region feature ($v_i$) extracted by the Bottom-up attention concatenated with
an embedding of previously generated word and the previous hidden state of the second LSTM that acts as a language model.
\par
After the hidden state of the attention LSTM is calculated based on input,
it is ready to be used for calculating the set of soft attention values $(\alpha)$.
the set of soft attention values are achieved via applying a SoftMax function over the attention values that come from the result of a non-linear transformation of
visual features from regions and the hidden state of the model at previous time step passed through linear transformation layers. 
\par
The set of attended visual features is calculated based on a soft-weight modulation of attention values over set of extracted visual features.
After the set of attended visual features is calculated it is ready to be concatenated with the current hidden state of attention LSTM
to form the input for language LSTM.
After the hidden state of language LSTM is created based on input, it is passed to a feed-forward SoftMax layer to create
the conditional distribution over possible output words.
\par
The Up-Down attention model has become a source of inspiration for most of the state-of-the-art image captioning models
as the quality of generated captions are superior to other models that do not use the bottom-up attention.
\par

\par
\subsection{Attention with Various Encoders \& Decoders}\label{sec4.1}
\par
In this section, we explore the state-of-the-art attentive models for image captioning that
have been published around the same time and after the advent of Up-Down attention model.
The reason is that Up-Down attention model produced competitive results
under all image captioning metrics compared to the models published
earlier, and therefore it can be considered as baseline for comparison with state-of-the-art models.
\subsubsection{CNN Encoders \& Attentive Decoders}\label{sec4.1.1}
\par
There are a few state-of-the-art models reviewed in this survey that use the traditional CNN encoder, only
using a CNN for feature extraction over the input image as whole.
Fig. \ref{fig3}
shows the general framework for models that employ CNNs for feature extraction alongside an attentive decoder for image captioning.
The Recurrent Fusion Network model proposed by Jiang et al. \cite{Jiang_ECCV_2018}, is
an example of using only a CNN backbone as the encoder part of the model and utilizing attentive LSTM as the decoder.
This model employs multiple CNN networks for feature extraction, where the extracted features are sent to multiple LSTMs
in the first fusion stage to create the soft multi-attention mechanism. The average of hidden states from LSTMs in the first fusion stage
is sent to LSTMs in the second fusion stage alongside multi-attention values. The attention values from each stage are concatenated to form the multi attention values.
The hidden states of LSTMs in the second fusion stage
are used in a soft-attention mechanism used in LSTM unit as decoder.
\par
Ye et al. \cite{Ye_2018_AttentiveLT}, employ a linear
transformation attention mechanism that applies soft attention on the features coming from the encoder.
This is done by using a feed-forward soft attention module that receives the extracted features from the CNN backbone,
followed by the application of the attended information to the hidden state of the attentive
LSTM decoder for additional information refinement.
Similarly, Chen et al. \cite{Chen_2018_ECCV} used two parallel linear transformations, joined with element-wise multiplication.
\par
The captioning model introduced by Shuster et al. \cite{Shuster_2019_CVPR} considers different personality categories
while generating captions, thus resulting in personalized captions that carry certain sentimental information.
For this purpose, a CNN (ResNext $32\times48d$  \cite{Resnext}) is used as image encoder and a
linear transformation of personality vector is used for encoding personality information.
In the decoder, LSTM is used like how it was utilized in Show and Tell, Show Attend and Tell, and the soft top-down decoder in Up-Down attention model as caption decoders.
They achieve the best results from the soft top-down attention decoder in Up-Down attention model.
The visual information is used in soft spatial attention like in Show, Attend and Tell.
The embedding of personality trait is used as additional input for LSTM in the decoder.
This model employs ResNet152 and ResNext$32\times48d$ as image encoders and thus the results of their experiments are not directly comparable to other methods
enumerated in Table \ref{table1}. As mentioned earlier, most of the state-of-the-art methods use ResNet101 or Faster-RCNN with ResNet101 as image encoder.
\par
Different from conventional soft attention mechanisms used with recurrent units, the Transformer \cite{NIPS2017_7181} model employs
scaled-dot (self) attention inside multi-head attention units eschewing recurrence and convolutions and instead fully relies on attention
for sequence generation.
\par
In this survey, among the models that use the Transformer model for image captioning,
the attentive model introduced by Zhu et al. \cite{Zhu_2019} is
the only one that uses a CNN as encoder.
The model introduced by Zhu et al. \cite{Zhu_2019}, employs the Transformer's decoder with a CNN followed by non-linear transformation. 
The CNN is used as the encoder instead of the original encoder in the Transformer \cite{NIPS2017_7181} model. The CNN backbones used for feature extraction
in this method are ResNet152 and ResNext101, making it hard to perform a direct comparison with results of other methods enumerated in Table \ref{table1}.
\par
All the methods mentioned above achieve good results under captioning metrics since they employ novel
ideas in terms of employing soft and multi-head attention in various new ways. The downside with these models is that they all rely on global features
extracted from the input image using a CNN and they do not leverage bottom-up attention, which provides object-level details.
\par
CNNs have also been used as language models rather than LSTMs. Aneja et al. \cite{Aneja_Conv} use a
soft attention mechanism on word embeddings of previously generated words and features from a convolutional encoder to form the attended
features for use in convolutional decoder. Although this model showed that convolutional networks can be used as a language model, the results
from this model are not competitive in comparison with the results from other state-of-the-art models.
\par
\subsubsection{Bottom-up Encoders}\label{sec4.1.2}
\begin{figure}[t!]
    \centering
    \includegraphics[width=\linewidth, frame={\fboxrule} {-\fboxrule}]{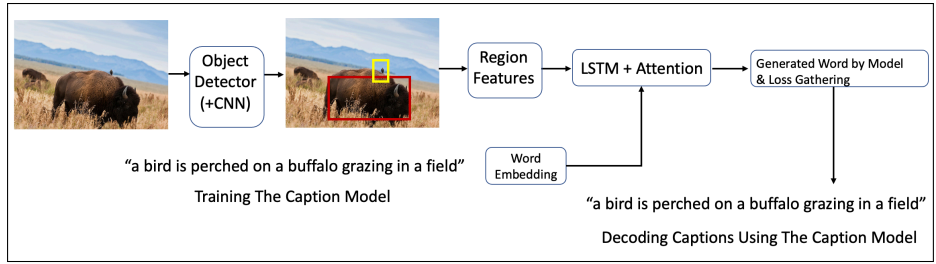}
    \caption{Attentive LSTM decoder with bottom-up attention encoder from a general point of view.}
    \label{fig6}
\end{figure}

Most of methods reviewed in our survey employ bottom-up attention \cite{Anderson2017up-down} as the encoder.
Fig. \ref{fig6} shows the general framework for models that employ bottom-up attention for feature
extraction alongside an attentive decoder for image captioning.
Previously, we discussed the relationship between the bottom-up and top-down
attention model introduced by Anderson et al. \cite{Anderson2017up-down}
and the attentive model used for visual grounding introduced by Lu et al. \cite{Lu2018NBT}.
Wang et al. \cite{Wang_2019_HAN} introduced a hierarchical soft attention module that considers
semantic information (concepts), bottom-up features and convolutional features extracted from equal sized patches in the image.
The attention values from different sources are combined with each other using a Multivariate Residual Module to be used by the language LSTM.
\par
Inspired by adaptive attention \cite{Lu2017KnowingWT}, Gao et al. \cite{Gao_Fan_etal_2019}
introduced the deliberate attention.
In deliberate attention network the previously generated hidden states of first and second residual LSTMs and global CNN
features along with an embedding of caption word are fed to the first residual LSTM, which acts as an attention LSTM in top-down attention module.
A linear transformation is applied to the concatenation of current hidden state of first residual LSTM and caption word embedding.
The results of this linear transformation layer is sent to the attention layer along with bottom-up features.
The attended visual features are sent to the second residual LSTM,
which acts as hierarchical attention LSTM, along with the results of linear transformation layer. 
Adaptive attention is then applied on the hidden state of second residual LSTM and regional features using a visual sentinel.
the visual sentinel is calculated based on linear transformations of the output of previous residual layer,
global features, attended regional features and previously generated hidden state of second residual LSTM.
A concatenation of the results of adaptive attention along with hidden state of second residual LSTM and linear transformation
of caption word embedding and hidden state of first residual LSTM are used in SoftMax function for next caption word generation.
\par
Ke et al. \cite{Ke_2019_ICCV}, employed
a bottom-up attention encoder with a reflective decoder.
The reflective decoder module includes a reflective attention module that employs soft-attention
and a reflective position module for alignment.
Qin et al. \cite{Qin_2019_CVPR} introduced an attentive model like the
bottom-up and top-down attention model \cite{Anderson2017up-down}. Instead of only considering the current attention values, the
attentive model proposed by Qin et al. \cite{Qin_2019_CVPR} also considers the attention values from previous iteration as well as the
attention values from current iteration.
Huang et al. \cite{Huang_2019_2} introduced the
adaptive attention time model that allowed the top-down attentive decoder to perform attention in the decoder multiple
times as adaptively as required by the model.
\par
\par
Yang et al. \cite{Yang_2019_ICCV} employ the idea of adding a control unit before
the language model LSTM. This control unit is used for selecting
the data source. At each time step, the model adaptively learns how much of the information about attributes,
visual features and relationships
must be used. Wang et al. \cite{Wang_Show_Recall_Tell} employ a recall unit that imitates
the way the human brain recalls previous experiences when
generating captions for images. This is done by adding a text-image matching unit.
The image features and textual features are embedded into a
common space, and the cosine similarity between them is calculated. This way a corpus of text
is generated for each image from the training data that includes the captions for each image. The recalled words are gathered from the corpus.
This allows the model to learn the best sentence structure.
\par
A novel Copy-LSTM unit was introduced by Sammani et al. \cite{Sammani_2020_CVPR}. Their attentive image captioning model titled as Show, Edit and Tell has a very similar
design to Up-Down attention model. They use the bottom-up attention features for a modified soft top-down attention decoder which includes an LSTM based denoising auto-encoder
along with the Edit-Net. The Edit-Net includes the top-down attention LSTM and Copy-LSTM which is replaced with the language LSTM in top-down attention module.
The Copy-LSTM unit allows the model to adaptively select and copy the memory state of the caption encoder LSTM with the highest soft attention values.
Rather than directly modulating the soft attention values on all context (memory) or hidden state as in conventional attention mechanisms, the
selective copy mechanism chooses the memory state with the highest value in the corresponding soft attention values.
This way the model is able to copy certain words from the input caption and edit the output caption based on the copied word.
\par
Zhou et al. \cite{Zhou_2020_CVPR} introduce a novel Part-of-Speech (POS) enhanced image-text matcher to act as a rewarder and an attention driver.
Soft attention mechanism is used in both the neural caption generator and POS enhanced image-text matching modules. A gated recurrent unit (GRU)
is used in the image-text matching module and the attention network is applied on the output of the GRU unit. In the neural caption
generator module, an attention network is applied to the visual and textual features. An LSTM is then used as a language model
to map the attended features to textual embeddings in a common space.
\par
At this time, bottom-up attention turns out to be the best kind of encoder used for image captioning as it considers object-level details and relationships.
However, models that employ only bottom-up attention as encoder do not consider semantic information and spatial relationships. For this reason,
graph convolutional networks (GCNs) have been
employed with bottom-up attention to provide semantic information and spatial relationships for the attentive language model.
\subsubsection{Bottom-up \& GCN Encoders}\label{sec4.1.3}
\par
\begin{figure}[ht!]
    \centering
    \includegraphics[width=\linewidth, frame={\fboxrule} {-\fboxrule}]{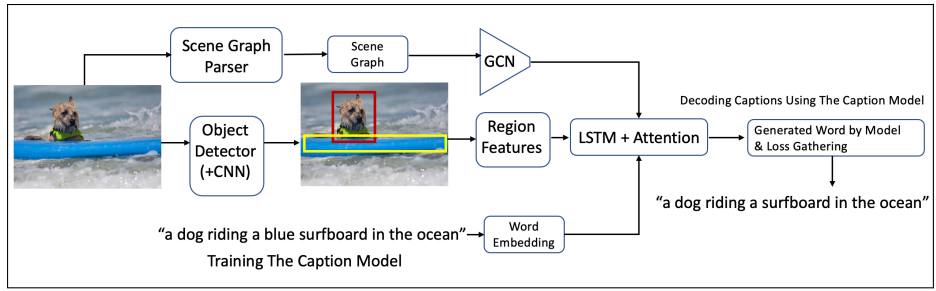}
    \caption{Attentive LSTM decoder with bottom-up attention and GCN as encoder from a general point of view.}
    \label{fig4}
\end{figure}

A new trend among the state-of-the-art methods is to employ bottom-up
attention and GCNs to form the attentive encoder. Fig. \ref{fig4}
shows the general framework for models that employ GCNs with bottom-up attention for image captioning. 
Inspired by how the SPICE \cite{SPICE} metric is calculated based on a scene graph generated from the input image, graph convolutional networks (GCNs) have been
used to extract information from scene graphs.
\par
\par
\par
Yao et al. \cite{Yao_2018_ECCV} employed bottom-up attention to provide the salient regions for use in two GCNs for spatial and semantic graphs, applying soft attention
on features extracted jointly from these graphs.
Wang et al. \cite{Wang_2019_role}, like the Up-Down attention model, used soft top-down attention, where the soft attention is applied
on features extracted from a spatial-semantic scene graph and image regions using the hidden state of attention LSTM.
Yao et al. \cite{Yao_2019_ICCV} use the Up-Down attention model and GCN-LSTM \cite{Yao_2018_ECCV} with a proposed hierarchical parsing module. 
The hierarchical parsing module leverages Faster-RCNN \cite{FasterRCNN} and Mask-RCNN \cite{Maskrcnn}
for detecting and segmenting the set of object regions and instances. 
The output of hierarchical parsing module is used
in a soft top-down attention to form the output of the model using the language LSTM unit.
There are also models that utilize scene graphs alongside bottom-up attention features without employing GCNs. For instance,
the attentive model introduced by Li \& Jiang \cite{Li_Jiang_2019_ieee} employs bottom-up attention
features and scene graph information represented in the form of semantic relationship features. The semantic
relationship features, and the bottom-up attention features are used to form the input to soft attention in the attentive decoder.
Yang et al. \cite{Yang_2019_CVPR} introduce a novel scene graph auto-encoder to be used in encoder-decoder model.
First, a scene graph auto-encoder is proposed that learns the shared dictionary, which includes the language inductive bias from sentence-to-sentence
reconstruction via employing scene graphs.
Then they proposed and used a multi-modal graph convolutional network
to re-encode visual features using a shared dictionary. The dictionary is shared between scene graph auto-encoder and re-encoder module in encoder-decoder
caption generator that receives visual features extracted via CNN. Like top-down attention in Up-Down model, soft attention is used for sentence reconstruction.
\par
Although adding scene graphs or GCNs to bottom-up attention successfully leverages spatial and semantic relationships
and achieves better performance compared with Up-Down attention model, these methods suffer from increased
model complexity in comparison with models that employ bottom-up features with multi-head attention in the Transformer model.
The attention calculation becomes more computationally expensive as the attention values are calculated based on semantic and spatial relationship features
alongside bottom-up features. As opposed to utilizing GCNs and scene graphs with bottom-up features for attention, using bottom-up features
in scaled-dot attention, which is a component of multi-head attention, achieves better performance and is computationally less expensive due to
parallelism in multi-head attention.
\par
\subsubsection{Bottom-up \& MHA Encoders}\label{sec4.1.4}
\par
\begin{figure}[ht!]
    \centering
    \includegraphics[width=\linewidth, frame={\fboxrule} {-\fboxrule}]{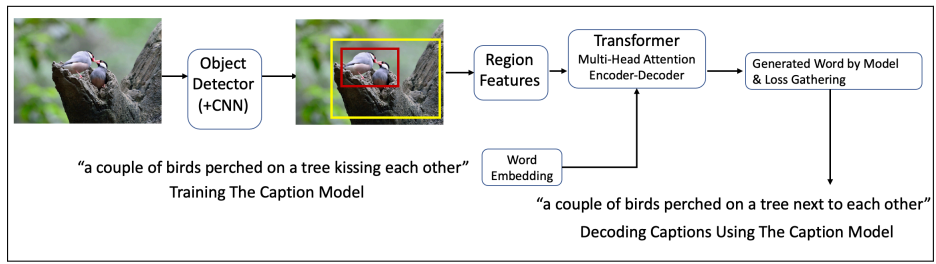}
    \caption{Attentive Transformer decoders with bottom-up attention encoder from a general point of view.}
    \label{fig5}
\end{figure}

Multi-head attention (MHA) that relies on multiple scaled-dot (self) attention heads, was shown to
be effective in the Transformer model \cite{NIPS2017_7181} for neural machine translation.
Multi-head attention was used to form both the encoder and decoder.
Fig. \ref{fig5} shows the general framework for models that employ multi-head attention to form the encoder and decoder for image captioning.
\par
\par
Herdade et al. \cite{Herade_2019_NIPS} used object names as attributes
to be added as input information for the attentive encoder. Yu et al. \cite{Yu_2019_mmt} introduced
an attentive model that employed multiple copies of
the object detector to provide the bottom-up attention encoder with various sets of object detections.
Liu et al. \cite{Liu_2019} similarly employ cross-modal information for this purpose.
Cornia et al. \cite{Cornia_2020_CVPR},
introduced the Meshed-Memory Transformer networks.
The Meshed-Memory Transformer networks are similar to the original Transformer networks,
except that the self-attention module is augmented with memory slots as plain learnable vectors that can be directly optimized using optimization
techniques such as stochastic gradient descent methods. Also, the memory states of each encoder in the encoder stack
is connected to all other decoders in the decoder stack using a mesh network connection followed by sigmoid activation.
Li et al. \cite{Li_2019_ICCV} introduced the 
Entangled Transformer encoder and used it for both visual and semantic information
to form the input for a multi-modal Transformer decoder. Instead of creating a mesh network connection between the memory states in Meshed-Memory
Transformer \cite{Cornia_2020_CVPR}, this model creates a mesh network connection between different queries at different time steps.
\par
\par 
Huang et al. \cite{Huang_2019_ICCV} introduced the Attention-on-Attention network (AoA-Net) model. In this model,
the encoder contains bottom-up attention and a refining module for bottom-up features, which includes multi-head attention followed by AoA module.
The decoder includes AoA module on top of an attention LSTM.
The AoA module applies a sigmoidal gate on the query concatenated with the output of lower attention module,
which could be multi-head attention or attentive LSTM, the result of concatenation is passed through a linear transformation before applying the sigmoidal gate.
The final result of the sigmoidal gate is multiplied by another linear transformation of the concatenation of the query and attention values.
Pan et al. \cite{Pan_2020_CVPR} improved the multi-head attention in Transformer model by adding a bilinear pooling mechanism to the self-attention used inside multi-head attention.
Using this mechanism, they introduced the x-linear attention block.
The x-linear attention leverages on both the spatial and channel-wise bilinear
attention values to extract the interactions between the input features \cite{Pan_2020_CVPR}.
The X-Linear attention block is used inside a multi-head attention module, which is placed on top of another x-linear attention block that receives the bottom-up features.
At each level, the output of x-linear blocks is embedded using linear transformation and the final output of multi-head attention with x-linear attention heads is sent to
the next x-linear attention block, which is placed on top of an LSTM. The output of LSTM and x-linear attention
are embedded using linear transformation and sent to a feed-forward SoftMax layer
for probability vector generation used for caption word generation at each time step.
In Table \ref{table1} We report the result of the experiments with x-linear attention used inside Transformer and multi-head attention with LSTM in x-linear attention network (X-LAN).
Guo et al. \cite{Guo_2020_CVPR} introduced the normalized self-attention and the geometry-aware
self-attention block that considers geometrical
information discovered from the visual features. Using a 4-dimensional vector containing the relative
position and the size of the bounding boxes for the objects,
the relative geometry features between objects are discovered.
\par
At the current time, the best results for image captioning come from models that employ scaled-dot and multi-head attention over bottom-up features and semantic information.
In comparison with models that employ GCN-LSTM architectures, multi-head attention-based methods achieve better performance, making them suitable for best practice
when employing attention mechanisms for image captioning.
\par

\subsection{Evaluation Metrics}
\par
In order to understand the quality of generated captions,
automatic metrics are used widely.
The survey by Liu et al. \cite{Liu2019_survey} offers useful details and information regarding how the evaluation metrics are designed.
Therefore, we avoid reiterating the same information here.
Without the need for manually verifying the quality of captions by a human agent,
automatic metrics such as BLEU \cite{BLEU_2002}, METEOR \cite{METEOR}, ROUGE \cite{ROUGE}, CIDER \cite{CIDER} and SPICE \cite{SPICE}
provide excellent insight regarding the quality
of captions, from different viewpoints. In order to perform a fair comparison of methods, we do not report the results under the ROUGE
metric as this metric is not commonly used in the majority of literature we discuss in Section \ref{sec4.3}.
\par
Originally designed for the evaluation of natural language generation tasks such as translation,
BLEU, METEOR and ROUGE,
provide numerical evaluation of the quality of translation from the input sequence to the output sequence based on n-gram overlap with ground truth translations.
Visual information is not considered when creating scores for these metrics and only the textual information regarding the input sequence 
and output sequence is considered.
\par
The BLEU metric counts and compares the number of occurrences of n-grams in ground truth (reference) caption and generated caption by the model.
The initial evaluation precision can be calculated by dividing the number of
n-grams in the generated caption that match the n-grams in the ground truth caption, by the length of generated caption (or translation).
However, here the translation recall rate is not considered for
certain meaningful and rare words, which leads to meaningless sentences achieving high precision score. To address this problem, the n-grams
are first counted in the generated caption sentence,
then the maximum number of n-grams is counted in each ground truth sentence.
The minimum value between the number of occurrences of n-grams in ground truth caption and generated caption by the model matching each other
and the maximum number of n-grams in each ground truth sentence is considered as the number of n-grams matching each other.
Because n-grams (usually between 1-4) are used for calculating the precision, therefore short sentence can highly affect the precision. To
address this problem a brevity penalty was introduced. The brevity penalty is set to one if the length of generated caption is longer than the length
of ground truth caption. If the length of ground truth caption is longer than the generated caption the brevity penalty is set to the exponential of
one minus length of generated caption divided by the length of ground truth caption.
Finally, BLEU is calculated via multiplying the brevity penalty to the exponential of average of log precision for different n-grams (1-4) resulting
in BLEU1, BLEU2, BLEU3 and BLEU4 metrics.
\par
The Meteor metric is calculated based on weighted average of single-precision rate and word recall rate.
In order to perform synonym and stem and word matching, Meteor calculates the value and recall rate of $Fmean$.
The average $Fmean$ of recall and precision accuracy
between the best candidate captions and ground truth captions is calculated via multiplying precision and recall divided by
a constant value (usually around 3) multiplied to precision plus one minus the same constant value multiplied to recall rate.
A penalty factor is introduced via multiplying a constant value (usually around 0.5) to
the square of number of chunks divided by the number of uni-grams.
Finally, the METEOR metric is calculated by multiplying one minus the penalty factor multiplied to the $Fmean$.
\par
To overcome the issue of not being able to consider visual
information for generating automatic metric scores,
Vedatanam et al. \cite{CIDER} introduced the CIDER score.
By employing term frequency and inverse document frequency (TF-IDF),
this metric evaluates the quality of a set of candidate sentences with
a set of ground truth sentences, given an input image. For the first time,
this metric provided a way to evaluate
the quality of generated captions considering fluency and relevancy;
fluency in caption sentence, and relevancy between 
the caption sentence and input image.
The CIDER score is achieved via calculating
the n-gram occurrences that match between the generated caption and ground truth captions.
Based on TF–IDF the n-grams are weighted and used for representing sentences in vector space.
The cosine similarity between the generated and ground truth sentence (TF-IDF) vectors is considered as CIDER score.
Although the CIDER metric was successful in assessing
relevancy between the images and sentences, there was no way to
verify the existence or truth regarding the presence of spatial relationships among the objects in the given image,
as described in the caption sentence generated by a model.
\par
To improve the quality of assessment, Anderson et al. introduced the SPICE metric, 
which employs scene graphs to verify the spatial
relationships among the objects, described in the caption.
In order to calculate the SPICE score, we need to embed the generated and ground truth captions
into an intermediate scene graph representation.
Via semantic parsing the captions are encoded into scene graphs.
The similarity score between the generated and ground truth captions scene graphs is calculated in the form of F1 score
from precision and recall. Two sets of logical tuples (that reflect semantic propositions) are created from each scene graph for generated and reference captions.
The matching tuples between logical tuples for generated and reference captions divided by the total number of logical tuples in
generated caption set is considered as precision and the number of matching tuples divided by total number of logical tuples in reference caption set
is considered as recall. Finally the F1 score (SPICE) is calculated via multiplying precision by recall by two divided by precision added to recall.
\par

\section{Discussion}\label{sec4.3.0}
\subsection{Comparison of attentive methods}\label{sec4.3}
\par
In this paper, we created categories for the attentive deep learning models used for image
captioning. Using these categories, we navigated
through important state-of-the-art models used for image
caption generation.
The comparison of performance of the methods categorized and reviewed for our survey is
shown in Tables \ref{table1}. The results reported in this table correspond to
experiments performed on the MS-COCO \cite{MS_COCO} dataset using Karpathy's test split \cite{Karpathy_2015_CVPR}. In this table, the first column
describes the reference for work explained in previous section. The second column explains the types of attention used in encoders and decoders.
The third column shows the kind of encoders and decoders. The results for experiments under CIDER and SPICE metrics are available in the fourth column and fifth column respectively.
The methods are first separated based on the type of attention used in decoders and second, they are
separated based on their performance under CIDER score.
\par
\begin{table}[!ht]\centering\label{table1}
    \caption{Comparison of performance among methods discussed in Section \ref{sec4.1}. Types of attention used in encoders and decoders
    (En-Att \& De-Att) could be bottom-up attention (BU), soft attention (SA), multi-head attention (MHA),
    or multi-head attention with bottom-up attention (BU+MH).
    Types of encoder could be (CNN) or object detector with CNN (OD), or a combination of these with (GCN) or Attention-on-Attention (AoA) \cite{Huang_2019_ICCV}.
    For decoder, we consider attentive (LSTM), (AoA), transformer (TR) or x-linear attention network (XLAN).
    For more technical details and differences regarding the methods and attention mechanisms please refer to Section \ref{sec4.1}.
    * indicates the results are reported using self-critical loss and also that the results of the same model trained with cross-entropy are available.
    $\dag$ indicates the results are reported only using self-critical loss. Other results are reported using cross-entropy loss.}
    \tiny
    \begin{tabular}{|l|c|c|c|c|c|c|c|}
        \toprule 
        \hline
        Reference & En-Att/De-Att & Encoder/Decoder & C & S & B4 & M \\
        \hline
        Pan et al 2020 \cite{Pan_2020_CVPR}* & BU+MH/MH & OD+TR/TR & \textbf{132.8} & \textbf{23.4} & \textbf{39.7} & \textbf{29.5} \\ 
        Li et al 2019 \cite{Li_2019_MDPI}$\dag$  & BU+HM/MH & OD+TR/TR & \textbf{131.5} & \textbf{23.2} & \textbf{39.7} & \textbf{29.4} \\
        Guo et al 2020 \cite{Guo_2020_CVPR}$\dag$  & BU+MH/MH & OD+TR/TR & \textbf{131.4} & \textbf{23.0} & \textbf{39.3} & \textbf{29.2} \\           
        Cornia et al 2020 \cite{Cornia_2020_CVPR}$\dag$ & BU+MH/MH & OD+TR/TR & \textbf{131.2} & \textbf{22.6} & \textbf{39.1} & \textbf{29.2} \\
        Yu et al 2019 \cite{Yu_2019_mmt}* & BU+MH/MH & OD+TR/TR & 130.9 & - & - & 29.1 \\
        Huang et al 2019 \cite{Huang_2019_ICCV}* & BU+MH/MH & OD+AoA/AoA & 129.8 & 22.4 & 38.9 & 29.2 \\         
        Liu et al 2019 \cite{Liu_2019}*  & BU+MH/MH & OD+TR/TR & 129.3 & 22.6 & 39.6 & 28.9 \\
        Herdade et al 2019 \cite{Herade_2019_NIPS}*  & BU+MH/MH & OD+TR/TR & 128.3 & 22.6 & 38.6 & 28.7 \\
        Li et al 2019 \cite{Li_2019_ICCV}* & BU+MH/MH & OD+TR/TR & 127.6 & 22.6 & 39.9 & 28.9 \\
        Pan et al 2020 \cite{Pan_2020_CVPR} & BU/MH & OD/XLAN & 122.0 & 21.9 & 38.2 & 28.2 \\ 
        Huang et al 2019 \cite{Huang_2019_ICCV} & BU+MH/MH & OD+AoA/AoA & 119.8 & 21.3 & 37.2 & - \\
        Li et al 2019 \cite{Li_2019_ICCV} & BU+MH/MH & OD+TR/TR & 119.3 & 21.6 & 37.8 & 28.4 \\         
        Yu et al 2019 \cite{Yu_2019_mmt} & BU+MH/MH & OD+TR/TR & 117.1 & - & 37.1 & 28.1 \\
        Liu et al 2019 \cite{Liu_2019}  & BU+MH/MH & OD+TR/TR & 118.2 & 21.2 & 37.9 & 28.3 \\
        Herdade et al 2019 \cite{Herade_2019_NIPS}  & BU+MH/MH & OD+TR/TR & 115.4 & 21.2 & 35.5 & 28.0 \\
        Yao et al 2019 \cite{Yao_2019_ICCV}*  & BU/SA & OD+GCN/LSTM & \textbf{130.6} & \textbf{22.3} & \textbf{39.1} & \textbf{28.9} \\
        Wang et al 2020 \cite{Wang_Show_Recall_Tell}* & BU/SA & OD/LSTM & \textbf{129.1} & \textbf{22.4} & \textbf{38.5} & \textbf{28.7} \\
        Sammani et al 2020 \cite{Sammani_2020_CVPR}* & BU/SA & OD/LSTM & \textbf{128.9} & \textbf{22.6} & \textbf{39.1} & - \\
        Yao et al 2018 \cite{Yao_2018_ECCV}*  & BU/SA & OD+GCN/LSTM & \textbf{128.7} & \textbf{22.1} & \textbf{38.3} & \textbf{28.6} \\
        Huang et al 2019 \cite{Huang_2019_2}*  & BU/SA & OD/LSTM & 128.6 & 22.2 & 38.7 & 28.6 \\         
        Yang et al 2019 \cite{Yang_2019_ICCV}* & BU/SA & OD/LSTM & 127.9 & 22.0 & 38.9 & 28.4 \\  
        Yang et al 2019 \cite{Yang_2019_CVPR}$\dag$ & BU/SA & CNN+GCN/LSTM & 127.8 & 22.1 & 38.4 & 28.4 \\
        Qin et al 2019 \cite{Qin_2019_CVPR}*  & BU/SA & OD/LSTM & 127.6 & 22.0 & 38.3 & 28.5 \\  
        Zhou et al 2020 \cite{Zhou_2020_CVPR}$\dag$  & BU/SA & OD/LSTM & 126.1 & 22.2 & 38.0 & 28.5 \\
        Gao et al. \cite{Gao_Fan_etal_2019}$\dag$ & BU/SA & OD/LSTM & 125.6 & 22.3 & 37.5 & 28.5 \\          
        Wang et al 2019 \cite{Wang_2019_HAN}*  & BU/SA & OD/LSTM & 121.7 & 21.5 & 37.6 & 27.8 \\         
        Yao et al 2019 \cite{Yao_2019_ICCV}  & BU/SA & OD+GCN/LSTM & 120.3 & 21.4 & 38.0 & 28.6 \\        
        Li \& Jiang 2019 \cite{Li_Jiang_2019_ieee}* & BU/SA & OD+GCN/LSTM & 120.2 & 21.4 & 36.3 & 27.6 \\ 
        Anderson et al 2018 \cite{Anderson2017up-down}* & BU/SA & OD/LSTM & 120.1 & 21.4 & 36.3 & 27.7 \\         
        Sammani et al 2020 \cite{Sammani_2020_CVPR} & BU/SA & OD/LSTM & 120.0 & 21.2 & 38.0 & - \\
        Huang et al 2019 \cite{Huang_2019_2}  & BU/SA & OD/LSTM & 117.2 & 21.2 & 37.0 & 28.1 \\
        Yao et al 2018 \cite{Yao_2018_ECCV}  & BU/SA & OD+GCN/LSTM & 117.1 & 21.1 & 37.1 & 28.1 \\
        Wang et al 2020 \cite{Wang_Show_Recall_Tell} & BU/SA & OD/LSTM & 116.9 & 21.3 & 36.6 & 28.0 \\
        Yang et al 2019 \cite{Yang_2019_ICCV}  & BU/SA & OD/LSTM & 116.6 & 20.8 & 37.1 & 27.9 \\ 
        Qin et al 2019 \cite{Qin_2019_CVPR}  & BU/SA & OD/LSTM & 116.4 & 21.2 & 37.4 & 28.1 \\
        Ke et al 2019 \cite{Ke_2019_ICCV} & BU/SA & OD/LSTM & 115.3 & 20.5 & 36.8 & 27.2 \\  
        Wang et al 2019 \cite{Wang_2019_HAN}  & BU/SA & OD/LSTM & 114.8 & 20.6 & 36.2 & 27.5 \\         
        Anderson et al 2018 \cite{Anderson2017up-down} & BU/SA & OD/LSTM & 113.5 & 20.3 & 36.2 & 27.0 \\
        Chen et al 2018 \cite{Chen_2018_ECCV}$\dag$  & -/SA & CNN/LSTM & 112.2 & - & 35.4 & 26.5 \\ 
        Ye et al 2018 \cite{Ye_2018_AttentiveLT} & -/SA & CNN/LSTM & 110.7 & 20.3 & 35.5 & 27.4 \\        
        Li \& Jiang 2019 \cite{Li_Jiang_2019_ieee} & BU/SA & OD+GCN/LSTM & 110.3 & 19.8 & 33.8 & 26.2 \\
        Wang et al 2019 \cite{Wang_2019_role}  & BU/SA & OD+GCN/LSTM & 108.6 & 20.3 & 34.5 & 26.8 \\                    
        Lu et al 2018 \cite{Lu2018NBT}  & BU/SA & OD/LSTM & 107.2 & 20.1 & 34.7 & 27.1 \\
        \hline
     \bottomrule 
    \end{tabular}
    \normalsize
\end{table}   
\par
The results reported for CIDER and SPICE metrics are borrowed from the original published literature,
based on experiments performed using cross-entropy loss. 
A self-critical loss was proposed by Rennie et al. \cite{selfcrit} to perform
reinforcement learning and optimization on the CIDER score directly.
While a few of the recently published literature reviewed in previous section report results using only one of these loss functions, the majority
report results using both loss functions.
Therefore, we show the results from experiments that use both loss functions.
We also ensure that ResNet101 or Faster-RCNN \cite{FasterRCNN} with ResNet101 \cite{Resnet} are commonly used for feature extraction among all compared methods to ensure fair comparison. 
Among all methods, in recurrent fusion network introduced by Jiang et al. \cite{Jiang_ECCV_2018} the network utilizes multiple different CNN networks for feature extraction.
Thus, the results of experiments via this model are not directly comparable with the results of other methods enumerated in Table \ref{table1}.
\par
The comparison of methods using automatic metrics such as CIDER and SPICE provides us with
insight about the quality of the generated captions in regards with visual features,
as well as insight regarding the effectiveness of different attention mechanisms employed by these methods.
\par
For conciseness, in Table \ref{table1} we show the performance of all the methods reviewed in Section \ref{sec4.1} in terms of CIDER and SPICE scores.
Although it is important to investigate results under other metrics, usually achieving higher scores under CIDER and SPICE metrics leads to higher scores under neural machine translation metrics
such as BLEU, METEOR and ROUGE. We also categorize different types of soft and multi-head attention mechanisms used in the methods discussed in Section \ref{sec4.1}
under soft attention (SA) and multi-head attention (MH).
The reason is we want to discover the effectiveness of employing variants of multi-head attention as opposed to variants of soft attention.
\par
In Section \ref{sec4.1}, we discussed the methods compared in Table \ref{table1} and categorized
them based on the way they employ attention mechanisms in their encoders.
Here we present all the methods reviewed in Section \ref{sec4.1}, based on the way they employ attention to create the attentive decoder.
We compare the methods based on performance using the scores
originally reported in the published literature.
By looking at Table \ref{table1} we realize that bottom-up attention with multi-head attention
encoders with multi-head attention decoders perform better than models
that employ variants of soft attention with LSTMs as decoder alongside bottom-up attention encoder.
\par
Another noteworthy attentive deep model used for image captioning offered by Cornia et al. \cite{Cornia_2019_CVPR}
employs the idea of controllability.
By creating different combinations for the order of the same detection set containing region proposals or
creating different detection sets with different objects, they create different control signals
that can generate different captions. The best control signal generates the caption with the highest CIDER score.
This mechanism leads to highly competitive results on the MSCOCO dataset, that are not comparable with the results gathered from
experiments with other attentive models for image captioning that do not employ controllability.
\par
Chen et al. \cite{Chen_shize_2020_CVPR} improve the idea of controllability
used in deep neural networks for image captioning by creating fine-grained control
signals that are selected based on the information gathered from scene graphs.
We did not bring the results from experiments that employ controllability into the table for comparison;
the reason is that the results from these experiments are not directly comparable with the results we have enumerated in Table \ref{table1}.
\par
We consider controllability as one the most important new concepts that can be used for boosting the quality of generated captions.
At the same time, recent work suggests that multi-head attention that employs scaled-dot attention, is more effective than soft attention.
Therefore, it makes sense to investigate the ways to create controllable and
grounded captions using multi-head attention and Transformer-based models.
\par
There are also other state-of-the-art deep image captioning models that are trained in semi-supervised fashion
without employing attention mechanisms. For instance, the dual generative adversarial network for image captioning by Liu et al. \cite{Liu_Dual_GAN}
does not employ attention mechanisms and instead relies on generative adversarial networks \cite{GAN_Goodfellow}.
\par
\subsection{Discussion of attentive methods}
\par
Visual soft and hard attention \cite{Xu} showed better performance in comparison
with earlier methods that did not benefit from application of attention mechanism
for image captioning \cite{Vinyals_2015_CVPR,Karpathy_2014,Karpathy_2015_CVPR}.
In comparison with bottom-up attention that employs object level features,
the downside of visual soft and hard attention is that they apply attention on visual
features from the whole image rather than specific regions in the image.
The soft attention was used over visual features from the whole image in most of the methods explained in Section \ref{sec4.0}
\par
Inspired by methods that applied attention on visual features towards object level features,
such as SCA-CNN \cite{Chen_2017_CVPR} and Areas of Attention \cite{Pedersoli_2017_ICCV}, bottom-up attention \cite{Anderson2017up-down}
employs an object detector with a CNN backbone that extract the visual spatial features from specific regions in the image. Bottom-up attention
improved the visual attention; however the attention was not utilized over semantic relationships among the detected regions in the image.
Bottom-up attention was used along various types of soft attention applied to visual bottom-up features with other sources of information as
explained in Section \ref{sec4.1.2}.
\par
Instead of utilizing soft attention over visual features, semantic attention \cite{You_2016_CVPR} leverages
soft attention on attributes discovered from the image using external resources.
Although the semantic attention achieved to better results in comparison with visual soft and hard attention, external resources should be used
for discovering attributes in the image, which leads to increased model complexity. The semantic attention became a source of inspiration for models
that employ GCNs (Section \ref{sec4.1.3}), which explore the spatial and semantic relationships between the regions of interest in the image. 
\par
Self-attention (scaled-dot attention) is a building block of multi-head attention. By applying self-attention on bottom-up attention features in the encoder and utilizing self-attention on 
the ground truth caption sequence while training, the model learns the semantic relationships among objects with reference to the semantic relationships
in the ground truth caption. This is mainly the reason why the methods explained in Section \ref{sec4.1.4}, which use multi-head attention with bottom-up attention
perform better than the methods explained in Section \ref{sec4.1.3}, which employ GCNs with bottom-up attention features.
\par
Limitations of all state-of-the-art methods explained in Section \ref{sec4.1} and the gaps among these methods are as the following.
The fact that bottom-up attention performs visual attention at object level improves the performance of all methods that employ this method, however a key question remains
whether in some wild scenarios the model needs to attend to regions in the image that do not contain any objects and rather contain natural or wild
elements that the object detector has not been trained to detect (such as mountain, trees, skies and etcetera).
\par
Regarding the self-attention in multi-head attention, because in the encoder of Transformer model the self-attention mechanism calculates
similarity scores between all elements of a sequence such as the set of bottom-up features from different regions in the image, therefore the question remains
if performing masking similar to how it is done in the decoder part of the Transformer model would benefit the model to learn more important relationships
among detected regions in the image rather than calculating the similarity scores between all regions.
\par
Although recently researchers have explored various methods for enhancing the multi-head attention and self-attention for image captioning as explained in
Section \ref{sec4.1.4}, an interesting question remains whether another attention component could be used rather than original self-attention which could potentially
leverage the bottom-up features or caption sequence in conjunction with bottom-up features better for caption generation.
The use of object detectors for extracting bottom-up features poses the risk of not being able to attend to regions that are relevant to captions in
domains that the object detector has no knowledge about (has not been trained on). Therefore, in wild scenarios more knowledge from various domains
might be required, which can be provided via object detectors trained on more specific tasks.
\par
\section{Conclusions}\label{sec5}
\par
We have performed a survey of state-of-the-art attentive deep learning models used for image captioning.
Comparison among the models reveals that Transformer-based models are leading
the way with highest scores for the image captioning metrics. Other methods
such as using bottom-up attention with soft-attention with LSTMs still work reasonably well and compete with multi-head attention and Transformer-based models.
Using Transformer-based models employing variants of multi-head attention with bottom-up attention results in better performance.
Investigating ways to improve the Transformers
or using them in innovative ways seems to be the right path for further research
related to neural attention mechanisms for image caption generation.
\par
Image captioning has become a major translation task in vision-language, combining the principles of computer vision and neural machine translation.
In the future, we are likely to witness widespread use of attention mechanisms in neural image captioning systems in various tasks.
Medical image captioning is an instance of a task which may be beneficial to the medical community.
Image captioning performed on mobile devices may become useful to visually impaired, especially when communicated verbally.
Bringing visual intelligence for robots and enhancing visual recommendation systems and visual assistant systems are among other applications of
neural image captioning. Attention mechanisms have shown the ability to map important parts of multi-modal information for use in downstream tasks
such as image captioning. Improving visual attention mechanisms (such as bottom-up, SCA-CNN, Areas-of-Attention) and multi-modal (soft) attention mechanisms as well as
multi-head attention and self-attention for image captioning remains an interesting challenge for the research community.
\par



%

\section*{Conflicts of interest/Competing interests}
The authors declare that they have no conflict of interest or competing interest in this work. N/A
\bibliographystyle{spmpsci}      
\bibliography{ALL.bib}   

%
%

\end{document}